\documentclass[runningheads]{llncs}

\usepackage{subfig}
\usepackage{graphicx}
\usepackage{amsfonts}
\usepackage{eucal}
\usepackage{amsmath}
\usepackage{booktabs}
\usepackage[algo2e]{algorithm2e}
\newcommand\norm[1]{\lVert#1\rVert}

\begin{document}
\titlerunning{ECGAN: Self-supervised GAN for electrocardiography}
\title{ECGAN: Self-supervised generative adversarial network for electrocardiography}
%
%
\author{Lorenzo Simone \inst{1}\orcidID{0000-0002-5010-7733} \and
Davide Bacciu\inst{1}\orcidID{0000-0001-5213-2468}}
\authorrunning{L. Simone, D. Bacciu}
%
\institute{Department of Computer Science, University of Pisa,  Pisa 56122, Italy}
\maketitle              
\begin{abstract}
High-quality synthetic data can support the development of effective predictive models for biomedical tasks, especially in rare diseases or when subject to compelling privacy constraints. These limitations, for instance, negatively impact open access to electrocardiography datasets about arrhythmias. This work introduces a self-supervised approach to the generation of synthetic electrocardiography time series which is shown to promote morphological plausibility. Our model (ECGAN) allows  conditioning the generative process for specific rhythm abnormalities, enhancing synchronization and diversity across samples with respect to literature models. A dedicated sample quality assessment framework is also defined, leveraging arrhythmia classifiers. The empirical results highlight a substantial improvement against state-of-the-art generative models for sequences and audio synthesis.


\end{abstract}
\section{Introduction}
\label{sec:intro}

Nowadays, the usage of electronic health records and the digital health (e-health) market have grown significantly due to technological advancements. Across the field, there is an upward trend in collecting heterogeneous data via diagnostic tests, clinical trials and wearable devices, supported by demographic insights. The recording of heart’s electrical activity (ECG), represents the most common non-invasive diagnostic tool for the early detection of life threatening conditions. 
Machine learning can play a key role in assisting clinicians by efficiently monitoring and stratifying the risk of patients (\cite{risk_strat}). Although fully supervised models could seem a flawless end-to-end solution for clinical research, they come with various drawbacks. For instance, such models are typically trained leveraging clinical datasets often failing in adequately covering rare occurrences. Furthermore, data annotation burden highly skilled physicians requiring exceptional time resources. Therefore, the quality of the predictions is strictly influenced by these preliminary phases, emphasizing the demand for plentiful rigorously labeled samples.


Deep generative models (DGMs), by approximating real conditional distributions, are usually capable of guiding the generative process towards a specific class of samples. For this reason, they represent a relevant tool for reducing unbalance from rare diseases via data augmentation or patient-specific synthesis (\cite{pat_spec}). The employment of DGMs as model agnostic synthesizers could implicitly solve another issue: patient’s
privacy and data anonymization. 
Potential threats arise from having a large amount of training data containing highly sensitive medical records. 

In this paper, we propose a novel architecture referred to as ECGAN, combining two fields of research: self-supervised learning (SSL) and deep generative models for time series. Our model has been devised specifically with  electrocardiography data in mind.  
The role of self-supervised learning in this research stands in exploiting the underlying time series dynamics via recurrent autoencoders. The features learned through a preliminary reconstruction task are transferred via weight sharing to the generator and a latent space projection. 
The assessment of the proposed model is compared with state of the art generative models for time series and audio synthesis (\cite{wavegan,crnngan,timegan}). It yields competitive results concerning structural fidelity, sampling diversity and data applicability to heart rhythm classification tasks. 

The main contributions of this work are: (1) We introduce a deep generative model, specifically designed for electrocardiography, intersecting between self-supervised learning and the generative adversarial framework. (2) We propose a parsimonious transfer learning framework requiring lesser resources than its generative counterparts. (3) We evaluate different  methods for quantitatively assessing generated ECGs, inspired by the evaluation of image generative models, we inspect the latent projection of an ECG arrhythmia classifier $\mathcal{C}$.
    


\section{Related work}
The complexity of modeling electrical heart's activity requires experts to cooperate across several fields of research. A pioneering research from \cite{rw7} theoretically investigated the application of dynamical systems to electrocardiography. The approach consisted in approximating heart's rhythm via a complex network of relaxation oscillators. This initiating idea spurred on upcoming methods built upon: non-linear coupled oscillators (\cite{rw8}); three-dimensional trajectories \cite{rw5,rw11} and function compositions \cite{rw13}. Those early efforts in dynamical models were able to produce highly realistic heartbeats, however they result oversimplified for augmentation procedures. 


Modern approaches often leverage generative deep learning, which sacrifices model explainability in favor of the avoidance of complex differential equations. One of the initial attempts from \cite{crnngan}, consisted in applying long-short term memories to continuous sequential music generation. The input to each cell is treated recurrently as a combination of noise and output from the previous time step. A similar conditional recurrent approach (RCGAN, \cite{rnngan}) replaced the association with previous time steps with conditional input embeddings. Notwithstanding the merit of those approaches, they merely focus on the adversarial minimax, without capitalizing on additional sources. On the other hand \cite{timegan} recently proposed TimeGAN, consisting of four components simultaneously learning to encode, generate, and iterate across time in an autoregressive procedure.

Recently, deep generative applications to electrocardiography have shown to be an emerging trend, mainly for improved heartbeat classification. Researchers have favourably adapted the promising image-based GAN framework to ECG settings (\cite{gan2}). At the same time, the latter have been combined with former ideas involving ordinary differential equations (ODEs) \cite{ode1,ode2}.

\section{The ECGAN Model}\label{sec:ecgan_model}
\begin{figure}[t]
\centering
\includegraphics[width=1.0\linewidth]{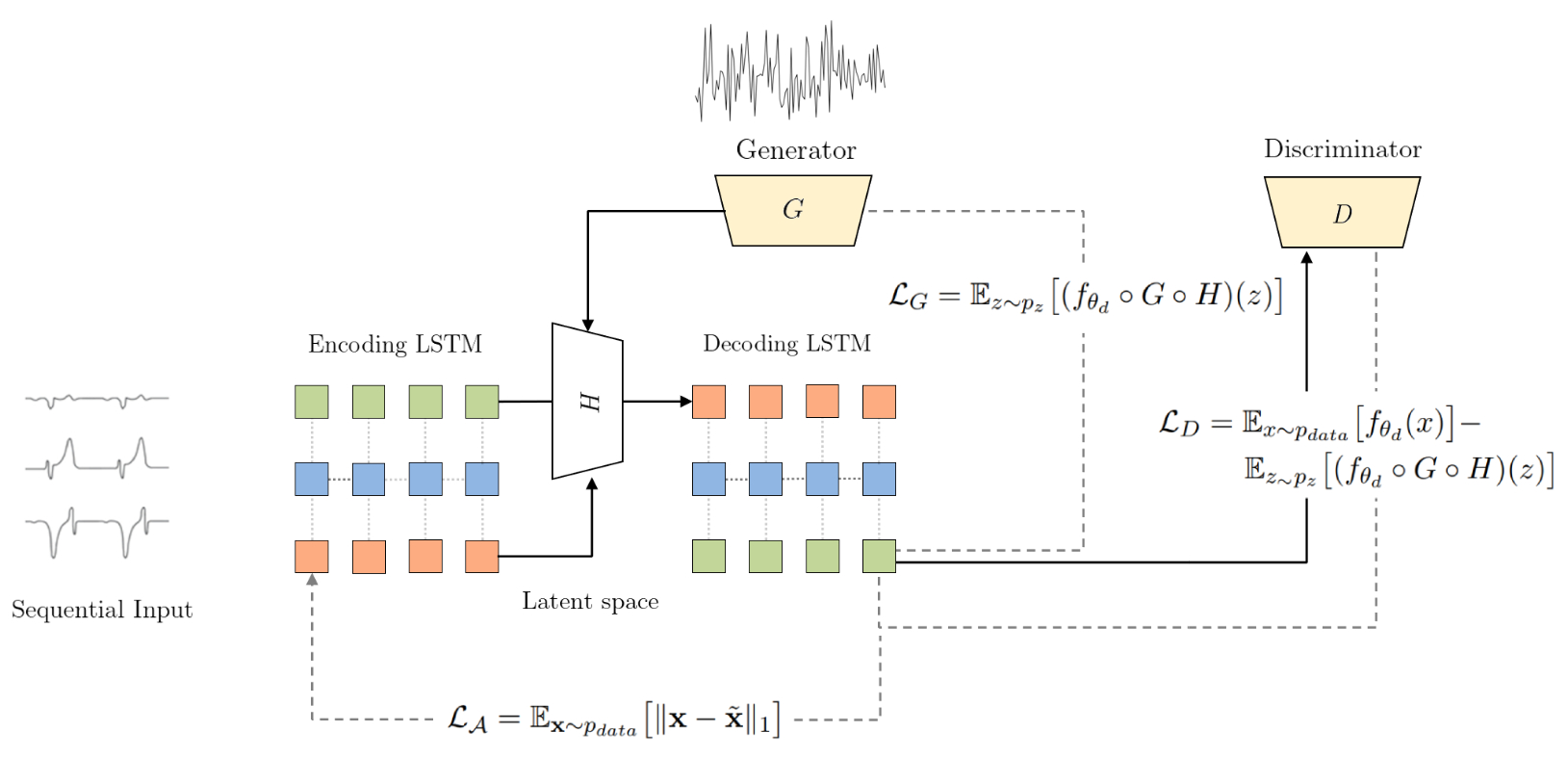}
\caption{Visual summary of the workflow of ECGAN. The main components (trapezoids) are correlated with their objective functions (dashed arrows). The sequential ECG input is either fed into the SSL block for preliminary reconstruction or to the discriminator $D$ for adversarial training. }
\label{fig:ecgan}
\end{figure}

\textbf{ECGAN Architecture}: The architecture encapsulates a latent space projection for the sequential domain within the adversarial framework typical of GANs. The model is composed of: an encoding recurrent mapping $E_\phi: \mathcal{X} \xrightarrow{}\mathcal{H}$, which produces a latent representation $\mathcal{H} \in \mathbb{R}^{h \times n'}$ through $H$ and a decoding recurrent block, sharing its weights with the generator $G(\mathcal{H}; \theta_g)$. We outlined each component in addition to their paths and training losses by dashed arrows in Figure \ref{fig:ecgan}.  There are two distinctive possible workflows, both traversing the latent space. The first path starts from the ECG input sequence and learns to produce faithful reconstructions without involving adversarial components. Finally, the generative process samples directly from the latent feature projection which is processed by $G$ and evaluated by D. 

Sequence generation is conditioned at global level using embedding layers. During training and sampling, the hidden state of the generator is initialized with a non-linear embedding of the corresponding ECG label. Likewise, the discriminator's output is perturbed by linearly combining its output and the associated embedding. Hence, they retain an initial conditional representation about the generated/classified sequence. This allows to both condition the generative process and influence the prediction based on prior class information.

\noindent\textbf{Training procedure}: The ECGAN training procedure comprises two independent phases. At an early stage, and differently from  standard GANs, the generator and the discriminator are kept in an idle state. Throughout this phase, unlabeled samples are used for a self-supervised reconstruction of the input. The main purpose is guiding the generative process towards the synthesis of partly recognizable ECGs instead of white noise. Finally, prior knowledge is transferred to the generator, henceforth sampling directly from the latent space projection. The full procedure summarising both phases is detailed in Algorithm \ref{alg:net}.

\begin{algorithm2e}[H]

\For{$t=0,1,\ldots epochs_{\operatorname{ssl}}$}{
        Sample $\{x^{(i)}\}_{i=1}^{m} \sim p_{data}$ \;
        $\delta_g \leftarrow \nabla_{\theta}[\frac{1}{m} \sum\limits_{i=1}^{m} \norm{{x^{(i)} - \tilde{x}^{(i)}}}_1]$ \;
        $\theta_g \leftarrow \theta_g-\alpha_s \cdot \mathrm{Adam}(\theta_g, \delta_g)$ \;
}

\For{$e=0,1,\ldots epochs_{\operatorname{adv}}$}{
	\For {$d=0,1,\ldots$steps}{
    	Sample $\{x^{(i)}\}_{i=1}^{m} \sim p_{data}$ \;
      Sample $\{z^{(i)}\}_{i=1}^{m} \sim p(z)$ \;
      $\delta_d \leftarrow \nabla_{\theta_d}[\frac{1}{m} \sum_{i=1}^{m} f_{\theta_d}(x^{(i)})-\frac{1}{m} \sum_{i=1}^{m} (f_{\theta_d} \circ G \circ H)(z)^{(i)}]$ \;
      $\theta_d \leftarrow \theta_d +\alpha_d \cdot$ RMSProp $(\theta_d, \delta_d)$ \;
      $\theta_d \leftarrow \operatorname{clip}(\theta_d,-c, c)$ \;
	}
		 Sample $\left\{z^{(i)}\right\}_{i=1}^{m} \sim p(z)$ \;
      	 $\delta_g \leftarrow-\nabla_{\theta} \frac{1}{m} \sum_{i=1}^{m}  
      	 (f_{\theta_d} \circ G \circ H)(z)^{(i)}$ \;
        	 $\theta_g \leftarrow \theta-\alpha_g \cdot \mathrm{RMSProp}(\theta_g, \delta_g)$ \;
}
\caption{Training procedure.}
\label{alg:net}
\end{algorithm2e}

The preliminary self-supervised phase is kept entirely modular depending on the downstream generative task. It requires an additional hyperparameter regulating the number of epochs and it employs Adam optimizer for the objective. Lastly, the adversarial training is carried out by using RMSProp for both the generator and the discriminator gradients. The discriminator is constantly kept within a compact parameter space through gradient clipping (an uniform window of [$-0.001, 0.001$] have been selected for all the experiments). Among possible combinations, the best working training ratio between the generator and the discriminator has been keeping it balanced ($g : d = 1 : 1$).

\noindent \textbf{Self-supervised module:} Training starts with a preliminary self-supervised task avoiding the usage of labeled patterns. We are solely interested in capturing high-level features through the manifold, exploiting them later for the downstream conditional generative task. It should be noted that the flexibility of the approach allows the extension to further SSL tasks such as signal denoising. The process can be summarized by the following three steps:
\begin{enumerate}
    \item Sampling a batch of samples from the real data distribution $\{x^{(i)}\}_{i=1}^{m} \sim p_{data}$
    \item Encoding the signal $x^{(i)}$ in a latent feature space through the embedding $E$ and the projection map $H$, thus producing $\mathcal{H} \in \mathbb{R}^{h \times n'}$.
    \item Recovering the original signal.
\end{enumerate}

Alternatively, we also investigated the effect of perturbing the signal with Additive White Gaussian Noise as an additional and more challenging SSL task. The perturbed signal is treated as $x'_j = x_j + z_j$ where $z_j \sim \mathcal{N}(0,\lambda^2 \sqrt{N})$ and we will refer to the corresponding model as $\operatorname{ECGAN}_\lambda$ throughout the experiments.

Then, we proceed by defining the reconstruction objective:

\begin{equation}
\label{eq:ssl}
\mathcal{L_A} = \mathbb{E}_{\mathbf{x} \sim p_{data}}{\big[\norm{\mathbf{x} - (G \circ H \circ E )(\mathbf{x})}_1\big]},
\end{equation}

\noindent while the gradient of the expectation is approximated through Montecarlo integration as 
$\nabla_{\theta}[\frac{1}{m} \sum\limits_{i=1}^{m} \norm{{x^{(i)} - (G \circ H \circ E )(\mathbf{x}^{(i)})}}_1]$, allowing us to train $G$, $H$ and $E$ through Adam optimizer \cite{adam}. Formally, the process requires an additional hyperparameter $s$, i.e. the number of training steps. Anyway, further investigations employing it in an alternating fashion among $G$ and $D$, are also suggested ($s\hspace{2pt}{:}\hspace{2pt}g\hspace{2pt}{:}\hspace{2pt}d = 0\hspace{2pt}{:}\hspace{2pt}1\hspace{2pt}{:}\hspace{2pt}1$ in the original GAN research). According to our experiments, the preliminary objective (\ref{eq:ssl}) as well as the self-supervised module introduced two desirable training properties:
\begin{itemize}
    \item The discriminator is inevitably forced to a rough start, being encouraged to look for finer distinctive features between real and fake samples (Figure \ref{fig:ssl}).
    \item Empirically, by inspecting pre-training checkpoints, the model grasps a primitive knowledge of ECG peaks location and duration (Figure \ref{fig:progression}).
        
\end{itemize}

\noindent \textbf{Adversarial Training:} Ultimately, the focus switches to the synchronized learning between the generator and the discriminator. Our preliminary experiments shared the widely known issues of standard GANs in encountering mode collapse scenarios (Appendix \ref{apd:modecollapse}). The adoption of combining WGAN principles introduced by \cite{wgan} and the proposed novelties heavily mitigated this phenomenon. Thus, the model minimizes the Wasserstein distance via the Kantorovich-Rubinstein duality: 

\begin{equation}\label{eq:wass_dist}
W(p_{data}, \tilde{p}) = \frac{1}{K} \sup_{\| f \|_L \leq K} \mathbb{E}_{x \sim p_{data}}[f(x)] - \mathbb{E}_{x \sim \tilde{p}}[f(x)].
\end{equation} 

\noindent Differently from the original research, we sample a noise vector $\mathbf{z} \sim p_z$ processing it through the pre-trained projection map $H$, which is subsequently fed to the generator. As a result, the objective for the generator is:

 \begin{equation}
\mathcal{L}_G = - \mathbb{E}_{z \sim p_z}{ \big[( f_{\theta_d} \circ G \circ H)(z) \big] }.
\end{equation}

\noindent Meanwhile, the discriminator is a real-valued mapping $f(\mathbf{x}; \theta_d) \colon \mathbb{R}^n \rightarrow \mathbb{R}$ using a linear activation function. Its K-Lipschitz continuity constraints in (\ref{eq:wass_dist}) are enforced by gradient clipping and the resulting objective is:

\begin{equation}
\mathcal{L}_D = \mathbb{E}_{x \sim p_{data}}{ \big[f_{\theta_d}(x)\big] } - \mathbb{E}_{z_\sim p_z}{ \big[( f_{\theta_d} \circ G \circ H)(z)\big] }.
\end{equation}

\noindent In other words, the discriminator should learn to estimate the distance between real patterns and synthesized ECGs, overcoming the prior advantage of the generator. Differently, the latter focuses exclusively on minimizing the Wasserstein distance between $p_{data}$ and $\tilde{p}$ in (\ref{eq:wass_dist}).

\section{Experimental Analysis}


Despite the growing interest in generative models for health, their evaluation is still predominantly qualitative. Other than being time-consuming and arduous to reproduce, this is highly subjective. Common metrics adopted for images such as the Inception Score (IS) based on the Inception Network (\cite{inception}) are inadequate for biomedical time-series. Since they are mainly pre-trained on datasets comprising objects from the real world, it discourages any fine-tuning on image-like signal representations.
Consequently, we pursued three core properties based on the application domain:
\begin{itemize}
    \item \textbf{Diversity}: Generated samples should be equally distributed among different classes of arrhythmias resembling the real data distribution
    \item \textbf{Fidelity}: The morphological properties of generated signals should mirror those from real samples (including segment lengths or complex duration).
    \item \textbf{Functionality}: New data should be adoptable and useful to enhance arrhythmia predictive models.
\end{itemize}

We propose an approach specifically designed for this sequential domain which includes an ECG arrhythmia classifier $\mathcal{C}$. The Inception feature space is replaced with an intermediate activation layer from the predictive model. The latter allows us to define $\phi(p_{data})$ and $\phi(\tilde{p})$ as Gaussian random variables with empirical means $\mu_{r}, \mu_{g}$ and covariances $\mathbf{C}_{r}, \mathbf{C}_{g}$. From now on, we rely on $ p_{\mathcal{C}} (y \mid \mathbf{x})$ as the label distribution of $\mathbf{x}$ predicted by $\mathcal{C}$ and $p_{\mathcal{C}}(y) = \int_{\mathbf{x}} {p_{\mathcal{C}}(y \mid \mathbf{x}) d \tilde{p}}$, being the marginal of $p_{\mathcal{C}}(y \mid \mathbf{x})$ over the generative probability distribution $\tilde{p}$. A detailed description and formulation of the metrics used throughout the experiments is covered in Appendix \ref{apd:first}.

To demonstrate the capability of the proposed model to synthesize realistic ECGs, we employed two different open datasets: MIT-BIH Arrhythmia Database and BIDMC Congestive Heart Failure Database. Each dataset is repeatedly split five times through a randomized hold-out technique, which assigns an equally partitioned 25\% dedicated to hyperparameters tuning and model assessment of the classifier $\mathcal{C}$. The remaining portion is entirely dedicated for the generative task (3998 and 11576 samples respectively).

\begin{figure}[t]
\centering
\captionsetup[subfigure]{labelformat=empty}
\caption{A T-distributed Stochastic Neighbor Embedding (t-SNE) visualization of the activations $\phi\left(p_{data}\right)$ and $\phi\left(\tilde{p}\right)$ obtained by the classifier $\mathcal{C}$. The distribution of synthesized samples from ECGAN (third column) and the top performing IS model (second column) is compared with the original datasets (first column).}
\subfloat{\label{fig:tsne-a}
  \hspace*{-1.2em}\includegraphics[width=0.3\linewidth]{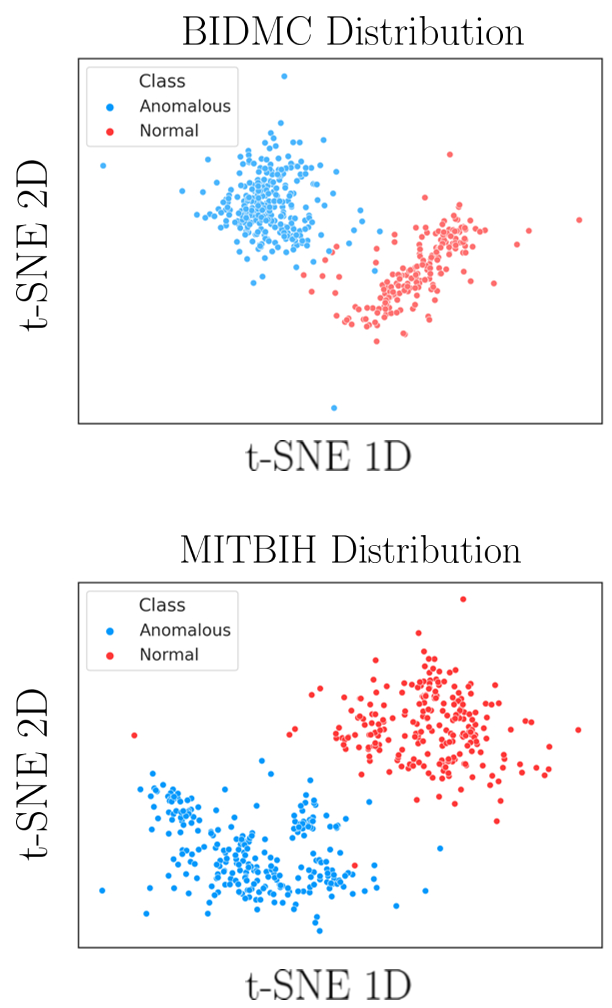}}
\hspace{25pt}
\subfloat{\label{fig:tsne-b}
  \hspace*{-1.2em}\includegraphics[width=0.3\linewidth]{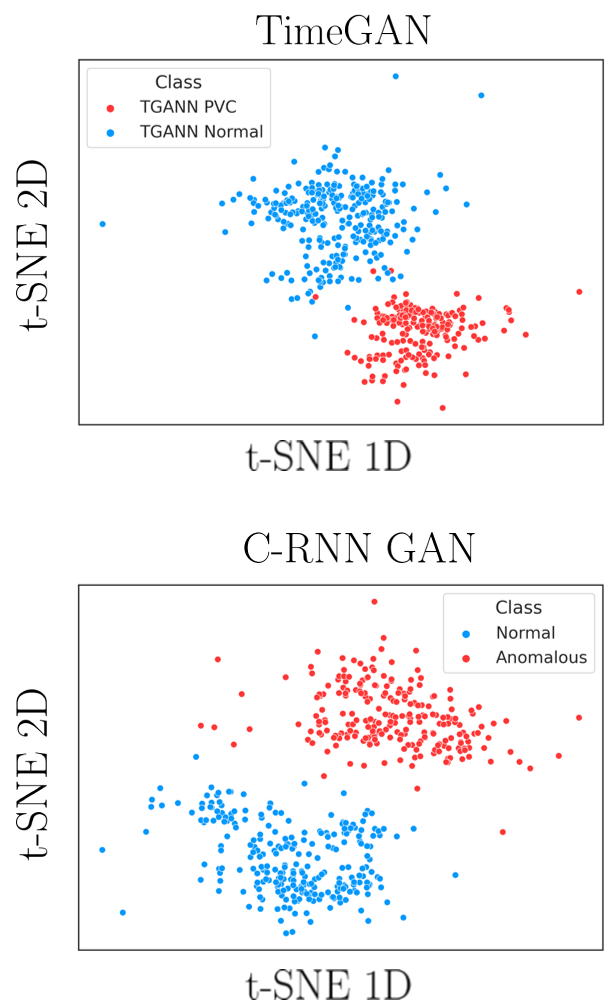}}
   \hspace{25pt}
  \subfloat{\label{fig:tsne-c}
  \hspace*{-1.2em}\includegraphics[width=0.3\linewidth]{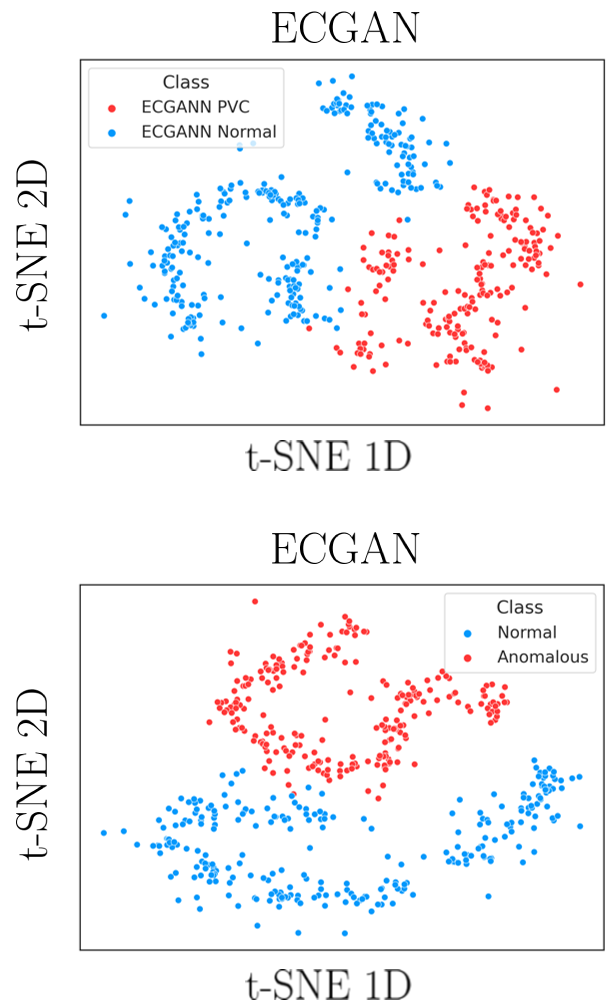}}
\label{fig:tsne}
\end{figure}

\begin{table}[!b]
\caption{Summary of the metrics evaluated for each generative model on both datasets. We report: the inception score (IS), the Frechèt Inception Distance (FID), the linear maximum-mean-discrepancy (MMD) and the Wasserstein distance (W).}
\begin{tabular}{   p{\dimexpr 0.2\linewidth-2\tabcolsep} 
                   p{\dimexpr 0.2\linewidth-2\tabcolsep} 
                   p{\dimexpr 0.2\linewidth-2\tabcolsep} 
                   p{\dimexpr 0.2\linewidth-2\tabcolsep} 
                   p{\dimexpr 0.2\linewidth-2\tabcolsep} }
\toprule
Model & $\text{FID}(p_{data}, \tilde{p})$ $\downarrow$ & $\text{IS}(\tilde{p})$ $\uparrow$ & $\text{MMD} (p_{data}, \tilde{p})$ $\downarrow$ & $\text{W} (p_{data}, \tilde{p})$ $\downarrow$\\
\midrule
 \multicolumn{5}{c}{BIDMC Dataset}\\
 \midrule
 ECGAN	 &	 $ \mathbf{233.86 \pm 17.24}$	 &	 $ \mathbf{1.97 \pm 0.09}$	 &	 $ \mathbf{31.71 \pm 4.43}$	 &	  $ \mathbf{0.52 \pm 0.01}$	\\
 $\operatorname{ECGAN}_{\lambda}$	 &	 $ 289.06 \pm 10.94$	 &	 $ 1.96 \pm 0.05$	 &	 $ 60.40 \pm 3.50$	 &	 $ 0.79 \pm 0.01$	\\
 TimeGAN	 &	 $ 328.03 \pm 2.13$	 &	 $ 1.60 \pm 0.01$	 &	 $ 144.90 \pm 3.51$	 &	 $ 1.03 \pm 0.00$	\\
 WaveGAN	 &	 $ 839.42 \pm 7.26$	 &	 $ 1.32 \pm 0.01$	 &	 $ 332.59 \pm 8.93$	 &
 $ 0.81 \pm 0.02$	\\
 C-RnnGAN	 &	 $ 917.22 \pm 63.54$	 &	 $ 1.35 \pm 0.03$ &	 $ 788.67 \pm 27.85$ &	 $ 2.07 \pm 0.03$	\\
 SpecGAN &	 $ 942.10 \pm 8.25$	 &	 $ 1.26 \pm 0.02$	 &	 $ 349.88 \pm 6.56$	 &
 $ 0.79 \pm 0.00$	\\
 
\bottomrule
\multicolumn{5}{c}{MITBIH Dataset}\\
\midrule
 ECGAN	 &	 $ \mathbf{45.49 \pm 1.82}$	 &	 $ \mathbf{1.41 \pm 0.03}$	 &	 $ \mathbf{17.41 \pm 1.62}$	& 	 $ 0.40 \pm 0.01$	\\
  $\operatorname{ECGAN}_{\lambda}$	 &	 $ 78.87 \pm 0.56$	 &	 $ 1.33 \pm 0.01$	 &	 $ 37.66 \pm 0.49$	 &	 $ 0.53 \pm 0.01$	\\
C-RNNGAN	 &	 $ 91.35 \pm 3.00$	 &	 $ 1.40 \pm 0.01$	 &	 $ 27.57 \pm 1.15$	 &	 $ 0.52 \pm 0.01$	\\
  TimeGAN	 &	 $ 113.81 \pm 2.37$	 &	 $ 1.11 \pm 0.01$	 &	 $ 60.51 \pm 2.23$	 &	 $ 0.68 \pm 0.01$	\\
  WaveGAN	 &	 $ 151.22 \pm 0.94$	 &	 $ 1.03 \pm 0.00$	 &	 $ 64.56 \pm 0.84$	 &	 $ \mathbf{0.32 \pm 0.00}$	\\
  SpecGAN	 &	 $ 151.21 \pm 0.95$	 &	 $ 1.01 \pm 0.00$	 &	 $ 64.97 \pm 0.58$	 &	 $ 0.35 \pm 0.01$	\\
\bottomrule
\end{tabular}
\label{tab:ext_results}
\end{table} 

The metrics collected throughout the experiments are compared with state of the art generative models for signals, including WaveGAN and SpecGAN \cite{wavegan}. These models were originally designed for audio synthesis, hence we performed a Short-time Fourier transform (STFT) to obtain spectrograms. We report the results over five different runs of the sampling procedure in Table \ref{tab:ext_results}. The proposed model consistently outperforms models selected from literature in both tasks considering the proposed metrics. Despite the increasing performance of the discriminative measures for the last task, the Inception Score suggests how spectrogram-based models are hardly producing well-distributed patterns. This is attributable to: an higher diversity of samples and the difficulty in masking zero padding from the original dataset.

\begin{figure}[t]
\captionsetup[subfigure]{labelformat=empty}
\centering

\includegraphics[width=0.9\linewidth]{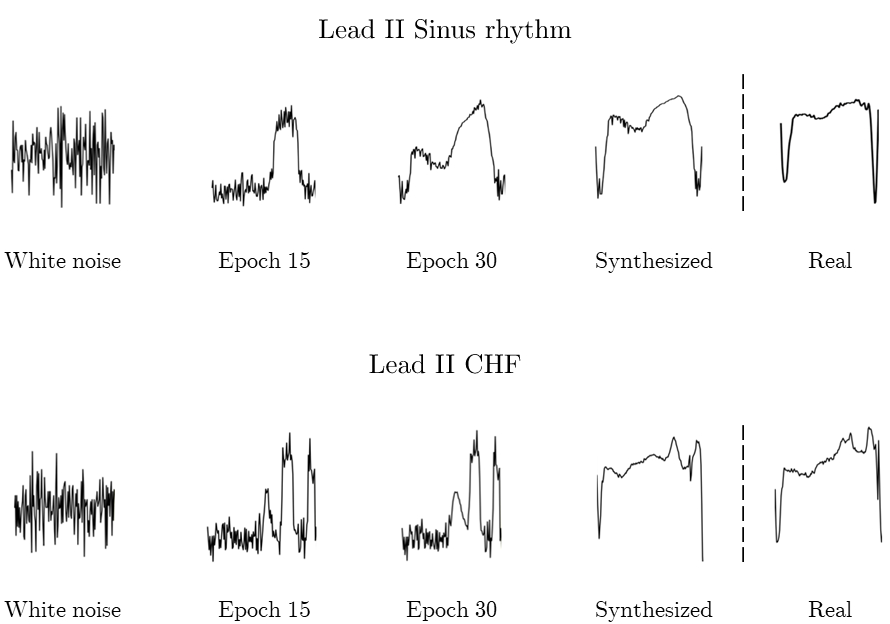}

\caption{Sampling progression over training epochs from ECGAN. The process starts with noise vectors and ends with a fully trained (SSL and adversarial) model sampling from sinus rhythm and congestive heart failure instances on the forth column.}
\label{fig:progression}
\end{figure}

In order to visually understand the capability of the classifier to distinguish synthesized patterns from original samples we computed a T-distributed Stochastic Neighbor Embedding (t-SNE \cite{tsnepaper}) of the activations $\phi(p_{data}), \phi(\tilde{p})$. The targeted real distribution in the first column exhibit two clearly distinguishable clusters for both datasets. On one hand, the competing models, according to the highest IS, follow the original separable arrangement (Figure \ref{fig:tsne}). On the other hand, ECGAN provides a unique spatial distribution of patterns. Hence, sample diversity across instances is maximized while maintaining a positional reference to the original classes. 

A contextual estimate of the impact of the SSL module and the subsequent adversarial phase over different training epochs is shown in Figure \ref{fig:progression}. We show, from left to right, the progression of a fixed random vector from the latent space through the SSL process. The fourth column represents the synthesized ECG pattern after the adversarial training. Lastly, the fifth column belongs to the best match, according to the DTW distance, between the generated and real pattern.

\section{Discussion}
We have introduced ECGAN, a generative model for ECG data leveraging self-supervised learning principles. Throughout the experiments, ECGAN has been shown to be able to produce morphologically plausible ECGs, including specific pattern abnormalities, by global conditioning of the input. 
Our self-supervised generative adversarial framework encourages the sampling process to synthesize patterns coherently to real samples' dynamics. Anyway, a proper trade-off between preliminary and adversarial phases by a rigorous hyperparameter selection is needed. Ultimately, we obtained competitive results for the expected properties over the monitored metrics. 

The synthesized patterns are also suitable for data augmentation procedures as suggested by the assessment of their functionality.
We believe that our contribution could pave the way to several other biomedical applications. As a first further development, we advise the usage of the proposed approach for multi-lead ECGs. This would potentially contribute with an heavier impact on improving deep-learning arrhythmia classifiers. A 12-lead ECG synthesizer could involve other recording systems such as vectorcardiography (VCG), which requires multiple leads. Encouraging trends emerge from investigating the explainability of generative biomedical models, besides the scope of improving predictive models. Furthermore, an additional promising application field for DGMs comes from electroencephalography (EEG). Similarly to our domain, data availability is inevitably affected by complex experimental setups and prolonged wearing sessions from the patients.

\bibliographystyle{splncs04}
\bibliography{bibliography}

\begin{thebibliography}{10}
\providecommand{\url}[1]{\texttt{#1}}
\providecommand{\urlprefix}{URL }
\providecommand{\doi}[1]{https://doi.org/#1}

\bibitem{wgan}
Arjovsky, M., Chintala, S., Bottou, L.: Wasserstein generative adversarial
  networks. In: Proceedings of the 34th International Conference on Machine
  Learning - Volume 70. p. 214–223. ICML'17, JMLR.org (2017)

\bibitem{risk_strat}
Beaulieu-Jones, B.K., Yuan, W., Brat, G.A., Beam, A.L., Weber, G., Ruffin, M.,
  Kohane, I.S.: Machine learning for patient risk stratification: standing on,
  or looking over, the shoulders of clinicians? NPJ digital medicine
  \textbf{4}(1), ~1--6 (2021)

\bibitem{wavegan}
Donahue, C., McAuley, J., Puckette, M.: Adversarial audio synthesis. In:
  International Conference on Learning Representations (2019)

\bibitem{rnngan}
Esteban, C., Hyland, S.L., R{\"a}tsch, G.: Real-valued (medical) time series
  generation with recurrent conditional gans. arXiv preprint arXiv:1706.02633
  (2017)

\bibitem{ode2}
Golany, T., Freedman, D., Radinsky, K.: Ecg ode-gan: Learning ordinary
  differential equations of ecg dynamics via generative adversarial learning.
  Proceedings of the AAAI Conference on Artificial Intelligence
  \textbf{35}(1),  134--141 (May 2021)

\bibitem{pat_spec}
Golany, T., Radinsky, K.: Pgans: Personalized generative adversarial networks
  for ecg synthesis to improve patient-specific deep ecg classification.
  Proceedings of the AAAI Conference on Artificial Intelligence
  \textbf{33}(01),  557--564 (Jul 2019). \doi{10.1609/aaai.v33i01.3301557}

\bibitem{ode1}
Golany, T., Radinsky, K., Freedman, D.: {S}im{GAN}s: Simulator-based generative
  adversarial networks for {ECG} synthesis to improve deep {ECG}
  classification. In: III, H.D., Singh, A. (eds.) Proceedings of the 37th
  International Conference on Machine Learning. Proceedings of Machine Learning
  Research, vol.~119, pp. 3597--3606. PMLR (13--18 Jul 2020)

\bibitem{adam}
Kingma, D.P., Ba, J.: Adam: A method for stochastic optimization. arXiv
  preprint arXiv:1412.6980  (2014)

\bibitem{tsnepaper}
Van~der Maaten, L., Hinton, G.: Visualizing data using t-sne. Journal of
  Machine Learning Research  \textbf{9},  2579--2605 (Nov 2008)

\bibitem{neurokit}
Makowski, D., Pham, T., Lau, Z.J., Brammer, J.C., Lespinasse, F., Pham, H.,
  Schölzel, C., Chen, S.H.A.: {NeuroKit}2: A python toolbox for
  neurophysiological signal processing. Behavior Research Methods
  \textbf{53}(4),  1689--1696 (feb 2021). \doi{10.3758/s13428-020-01516-y}

\bibitem{rw5}
McSharry, P.E., Clifford, G.D., Tarassenko, L., Smith, L.A.: A dynamical model
  for generating synthetic electrocardiogram signals. IEEE transactions on
  biomedical engineering  \textbf{50}(3),  289--294 (2003)

\bibitem{crnngan}
Mogren, O.: C-rnn-gan: A continuous recurrent neural network with adversarial
  training. In: Constructive Machine Learning Workshop (CML) at NIPS 2016. p.~1
  (2016)

\bibitem{rw13}
Roonizi, E.K., Sameni, R.: Morphological modeling of cardiac signals based on
  signal decomposition. Computers in biology and medicine  \textbf{43}(10),
  1453--1461 (2013)

\bibitem{rw11}
Sameni, R., Clifford, G.D., Jutten, C., Shamsollahi, M.B.: Multichannel ecg and
  noise modeling: Application to maternal and fetal ecg signals. EURASIP
  Journal on Advances in Signal Processing  \textbf{2007},  1--14 (2007)

\bibitem{rw8}
dos Santos, A.M., Lopes, S.R., Viana, R.R.L.: Rhythm synchronization and
  chaotic modulation of coupled van der pol oscillators in a model for the
  heartbeat. Physica A: Statistical Mechanics and its Applications
  \textbf{338}(3-4),  335--355 (2004)

\bibitem{gan2}
Sarkar, P., Etemad, A.: Cardiogan: Attentive generative adversarial network
  with dual discriminators for synthesis of ecg from ppg. Proceedings of the
  AAAI Conference on Artificial Intelligence  \textbf{35}(1),  488--496 (May
  2021)

\bibitem{inception}
Szegedy, C., Ioffe, S., Vanhoucke, V., Alemi, A.A.: Inception-v4,
  inception-resnet and the impact of residual connections on learning. In:
  Thirty-first AAAI conference on artificial intelligence (2017)

\bibitem{rw7}
Van Der~Pol, B., Van Der~Mark, J.: Lxxii. the heartbeat considered as a
  relaxation oscillation, and an electrical model of the heart. The London,
  Edinburgh, and Dublin Philosophical Magazine and Journal of Science
  \textbf{6}(38),  763--775 (1928)

\bibitem{timegan}
Yoon, J., Jarrett, D., Van~der Schaar, M.: Time-series generative adversarial
  networks. Advances in neural information processing systems  \textbf{32}
  (2019)

\end{thebibliography}

\appendix
\newpage
\section{Dataset Preparation}\label{sec:data_prep}

We considered two ECG classification datasets, both publicly accessible from PhysioNet. The first dataset have been used as a baseline, since it has a lower diversity across patterns and it has been automatically annotated.

\noindent \textbf{BIDMC Congestive Heart Failure Database (ECG5000):} 

\noindent The open-source dataset \textit{BIDMC-CHFDB} is a collection of long-term ECG recordings from 15 subjects suffering from severe congestive heart failure (NYHA class 3-4). The initial recordings were twenty hours long, and comprised two-lead ECG signals sampled at 250 Hz with a 12-bit resolution over a range of $\pm10 mV$. They have been collected at Boston's Beth Israel Hospital with an approximate  bandwidth of $0.1-40\text{Hz}$. The collection is composed of normal rhythm beats as well as premature atrial and ventricular complexes (PVC) and R-on-T PVCs.

\noindent \textbf{Preliminary procedures:} The first dataset, being patient-specific, serves as a baseline task for the generative models examined. Five thousand singular beats have been extracted following the principles adopted for the dataset "ECG5000". After having centered each fragment by its QRS complex, we obtained sequences of equal length by linear interpolation. 


\noindent \textbf{MIT-BIH Arrhythmia Database:} 

\noindent The MIT-BIH Arrhythmia Database is composed of 48 half-hour two-lead ambulatory ECG recordings. It comprises 47 different subjects and it was collected at the BIH Arrhythmia Laboratory.
For the generative task, we have used the II lead undersampling it from 360Hz to 125Hz. Differently from the first dataset, it provides five annotated classes performed by two cardiologists ( in accordance with the Association for the Advancement of Medical Instrumentation AAMI-EC57.

\noindent \textbf{Preliminary procedures:}
The initial distribution of samples across the five classes is highly unbalanced. We proceeded by discarding the classes \textit{S} and \textit{F}, the first being a collection of several atrial and supra-ventricular events and the second being a mixture of the latter. This could unnecessarily complicate the generative task. The resulting classes: \textit{N} referring to normal rhythm and \textit{V} standing for premature ventricular complexes were highly unbalanced (\textit{N} contains 72470 samples, while \textit{V} only 5788). In order to balance the two classes for the generative task, we performed a standard undersampling procedure.

\section{Metrics definition}\label{apd:first}

The discrepancy measures from the experimental results are detailed utilizing the label distribution $p_{\mathcal{C}} (y | \mathbf{x})$ and its marginal distribution $p_{\mathcal{C}}(y)$ from $\mathcal{C}$.
\vspace{5pt}\\
The Inception Score (IS) is computed as:
\begin{equation}
\operatorname{IS}(\tilde{p})=e^{\mathbb{E}_{\mathbf{x} \sim \tilde{p}}[K L(p_{\mathcal{C}}(y \mid \mathbf{x}) \| p_{\mathcal{C}}(y))]}    
\end{equation}

The Fréchet Inception Distance (FID) is:

\begin{equation}
\operatorname{FID}(p_{data}, \tilde{p})=\|\mu_{r}-\mu_{g}\|+\operatorname{Tr}(\mathbf{C}_{r}+\mathbf{C}_{g}-2(\mathbf{C}_{r} \mathbf{C}_{g})^{1 / 2})
\end{equation}

Kernel-based Maximum Mean Discrepancy (MMD) used as a dissimilarity measure between $p_{data}$ and $\tilde{p}$:

\begin{equation}
\operatorname{MMD}^{2}(p_{data}, \tilde{p})=\underset{\mathbf{x}_{r}, \mathbf{x}_{r}^{\prime} \sim p_{data},}{\mathbf{x}_{g}, \mathbf{x}_{g}^{\prime} \sim \tilde{p}}[k(\mathbf{x}_{r}, \mathbf{x}_{r}^{\prime}) -2 k(\mathbf{x}_{r}, \mathbf{x}_{g})+k(\mathbf{x}_{g}, \mathbf{x}_{g}^{\prime})]
\end{equation}

\noindent ,with a linear and radial basis function (RBF) kernel $k$:
\begin{equation}
k(\mathbf{x}_{r}, \mathbf{x}_{g}) = exp(-\frac{||\mathbf{x}_{r}, \mathbf{x}_{g}||^2}{2\sigma^2})
\end{equation}

\section{Ablation study}\label{apd:ssl_effect}

The focus in this appendix section is devoted to the empirical estimation of the contribution of the proposed novelties. We start by singularly detaching each component analyzing the impact on the evaluated metrics, following with qualitative improvements over adversarial training stability and finally presenting the latter over the generative process. We will refer, from now on, to a fully adversarial version of ECGAN having the self-supervised component and training removed as $\operatorname{ECGAN}_{\operatorname{NO-SSL}}$. 
The already mentioned set of metrics have been reused for the experiments in Table \ref{tab:abl_results}. The main difficulties without self-supervision have been: producing high quality ECGs, avoiding mode collapse and a lack of improvement over predictive metrics for classification tasks.

\begin{table}[!h]
\caption{Summary of the metrics for the ablation study of the SSL module evaluated on both datasets.}
\begin{tabular}{   p{\dimexpr 0.2\linewidth-2\tabcolsep} 
                   p{\dimexpr 0.2\linewidth-2\tabcolsep} 
                   p{\dimexpr 0.2\linewidth-2\tabcolsep} 
                   p{\dimexpr 0.2\linewidth-2\tabcolsep} 
                   p{\dimexpr 0.2\linewidth-2\tabcolsep} }
\toprule
Configuration & $\text{FID}(p_{data}, \tilde{p})$ $\downarrow$ & $\text{IS}(\tilde{p})$ $\uparrow$ & $\text{MMD} (p_{data}, \tilde{p})$ $\downarrow$ & $\text{W} (p_{data}, \tilde{p})$ $\downarrow$\\
\midrule
 \multicolumn{5}{c}{BIDMC Dataset}\\

\midrule
$\operatorname{ECGAN}$	 &	 $ 233.86 \pm 17.24$	 &	 $ 1.97 \pm 0.09$	 &	 $ 31.71 \pm 4.43$	 &	  $ 0.52 \pm 0.01$	\\
  $\operatorname{ECGAN}_{\operatorname{NO-SSL}}$	 &	 $ 1156.0 \pm 13.83$	 &	 $ 1.03 \pm 0.02$	 &	 $ 728.91 \pm 13.95$	 &	 $ 1.19 \pm 0.01$	\\

 \midrule
\multicolumn{5}{c}{MITBIH Dataset}\\
\midrule
$ \operatorname{ECGAN}$	 &	 $ 45.49 \pm 1.82$	 &	 $ 1.41 \pm 0.03$	 &	 $ 17.41 \pm 1.62$	 &	 $ 0.40 \pm 0.01$	\\
  $\operatorname{ECGAN}_{\operatorname{NO-SSL}}$	 &	 $ 1189.46 \pm 12.83$	 &	 $ 1.02 \pm 0.10$	 &	 $ 752.1 \pm 8.45$	 &	 $ 1.20 \pm 0.01$	\\
\bottomrule
\end{tabular}
\label{tab:abl_results}
\end{table}

\begin{figure}[h]
\centering
\captionsetup[subfigure]{labelformat=empty}
\subfloat[]{\includegraphics[width=0.4\textwidth]{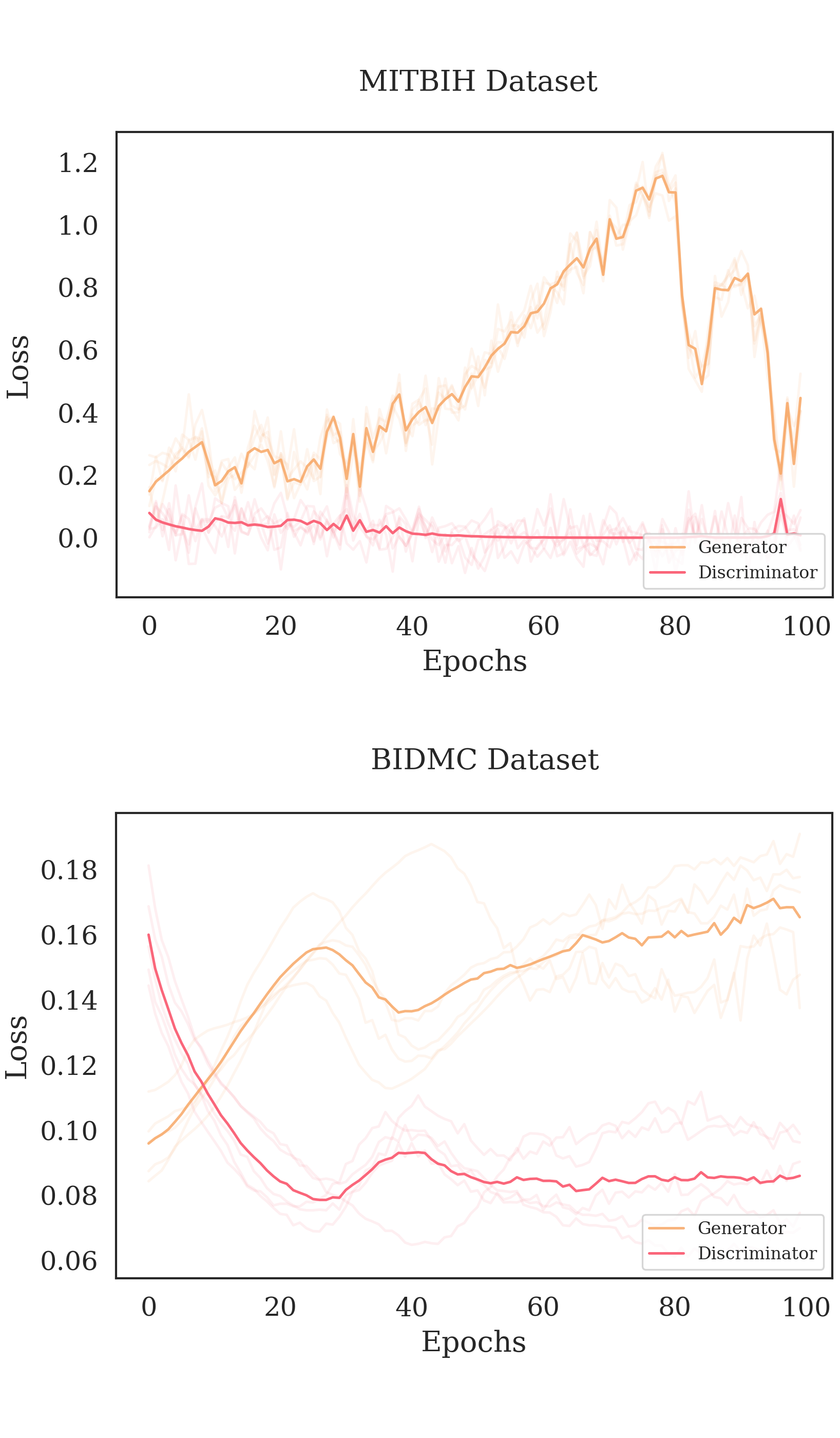}\label{fig:ssl-a}}
\hspace{45pt}
\subfloat[]{\includegraphics[width=0.4\textwidth]{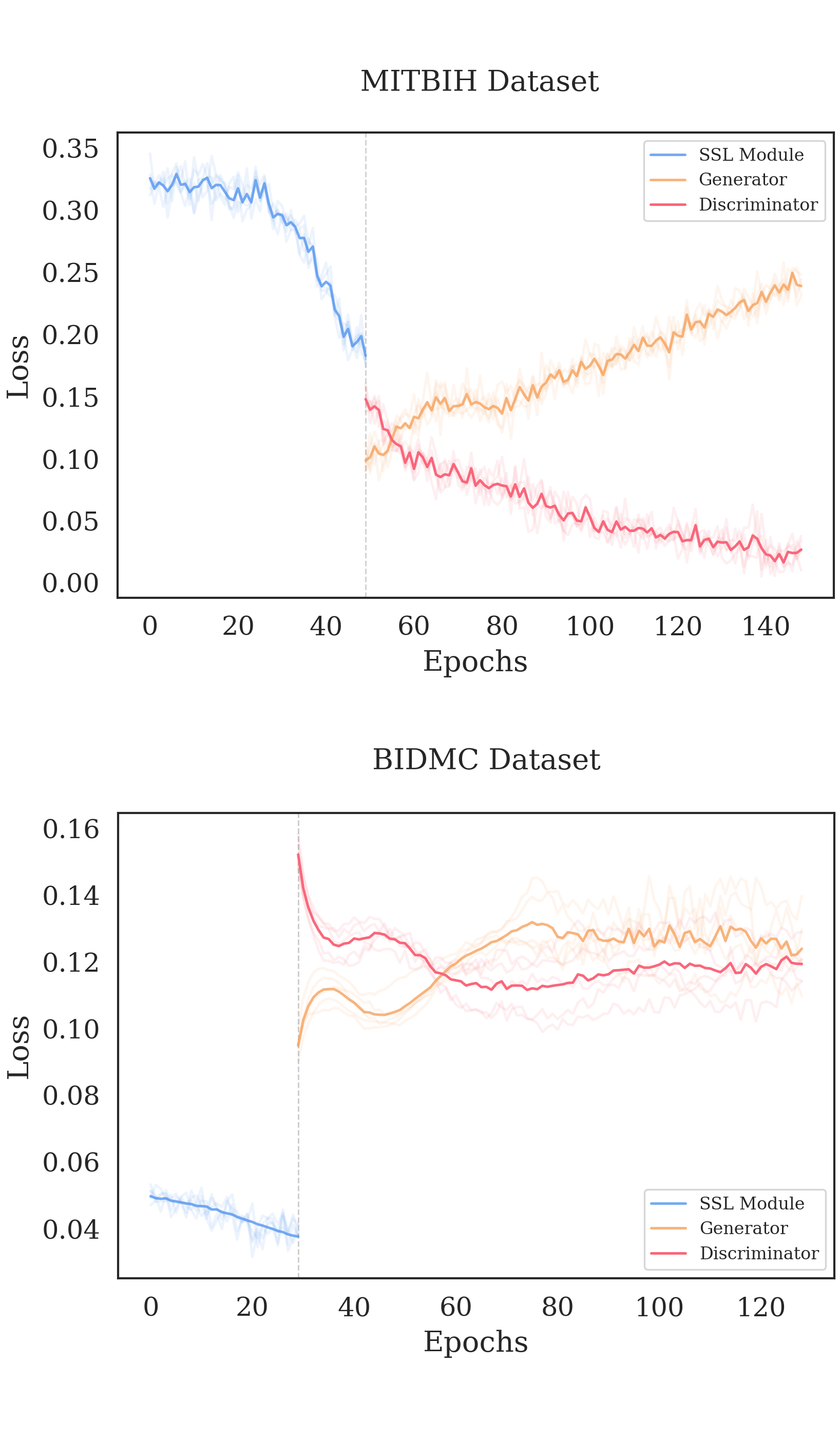}\label{fig:ssl-b}}
\caption{The loss over training epochs of the generator and the discriminator from
standard GAN (first column) is compared with its counterparts and the SSL module from
the proposed model (second column).}
\label{fig:ssl}
\end{figure}

Throughout the experiments we have monitored the loss of inner components from ECGAN, namely the generator, the discriminator and the SSL module as shown in Figure \ref{fig:ssl}. In the first column, over five different runs, the discriminator starts to improve significantly at an early stage. As a consequence, there is a clear downward trend for the generator, which is incapable of improving nor synthesize plausible samples. The pre-training SSL module contributes to mitigating this phenomenon promoting a balance between $G$ and $D$.

\section{Mode collapse}\label{apd:modecollapse}
\begin{figure}[!htb]
\centering
\subfloat[]{\label{fig:trainmc-a}
  \includegraphics[width=0.30\linewidth]{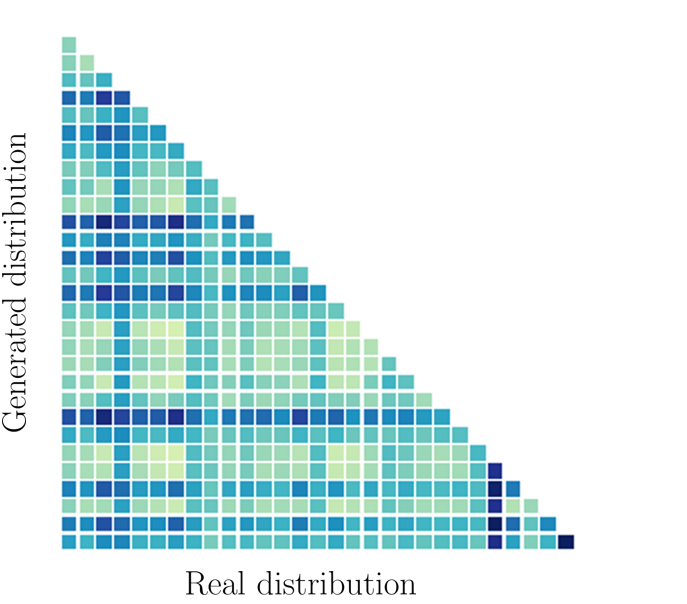}}
\hspace{40pt}
\subfloat[]{\label{fig:trainmc-b}
  \includegraphics[width=0.30\linewidth]{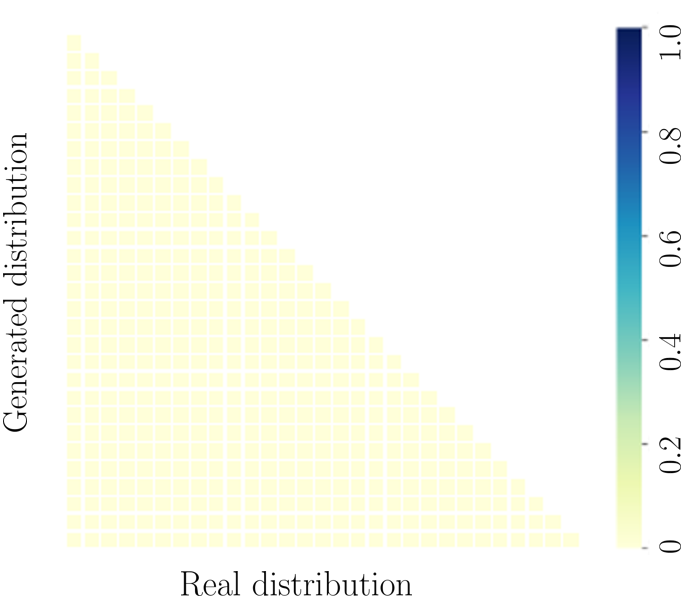}}
\caption{DTW Correlation matrix over 30 samples from ECGAN (a) and standard GAN (b), which is suffering from mode collapse. }
\label{fig:mode_collapse}
\end{figure}

Standard generative adversarial networks have been proven their strengths, especially in image generative tasks, over the last years of research. Still, they pose several difficulties in having a balanced learning process. We reported the limitations encountered by training them from both qualitative and quantitative perspectives. The latter have been measured computing the dynamic time warping (DTW) distance over 30 generated patterns from ECGAN and standard GAN (Figure \ref{fig:mode_collapse}).

\begin{figure}[!htb]
\centering
\caption{Synthesized sample from standard GAN (b) failing in learning a reliable approximation of the real data distribution. There is a clear evidence of \textit{mode collapse} with a fair discrepancy from real patterns (a).}
\subfloat[]{\label{fig:mc_signal-a}
  \includegraphics[width=0.3\linewidth]{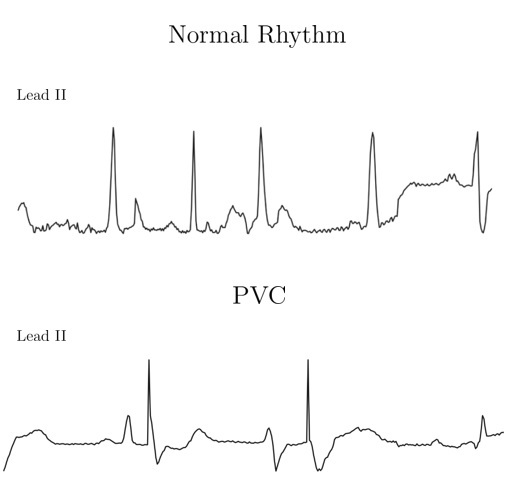}}
\hspace{60pt}
\subfloat[]{\label{fig:mc_signal-b}
  \includegraphics[width=0.3\linewidth]{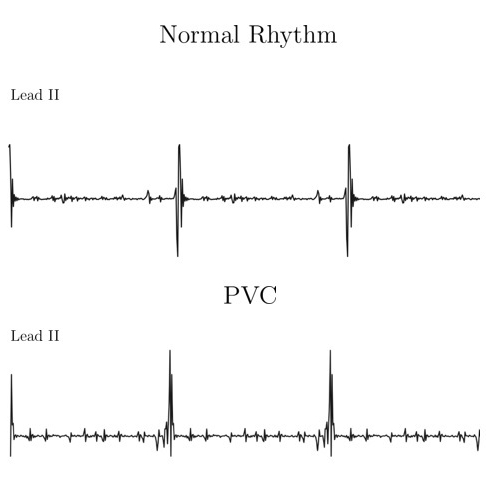}}
\label{fig:mc_signal}
\end{figure}

The overall quality of the generative process is unsatisfactory other than suffering from a noticeable mode collapse behaviour (Figure \ref{fig:mc_signal}). Apart from being capable of producing R peaks, there are not other distinctive features in the produced signal (Figure \ref{fig:mc_signal-b}).

\newpage
\section{Segments analysis}\label{apd:segments}

\begin{figure}[!htb]
\centering
\captionsetup[subfigure]{labelformat=empty}
\subfloat[]{\includegraphics[width=0.35\textwidth]{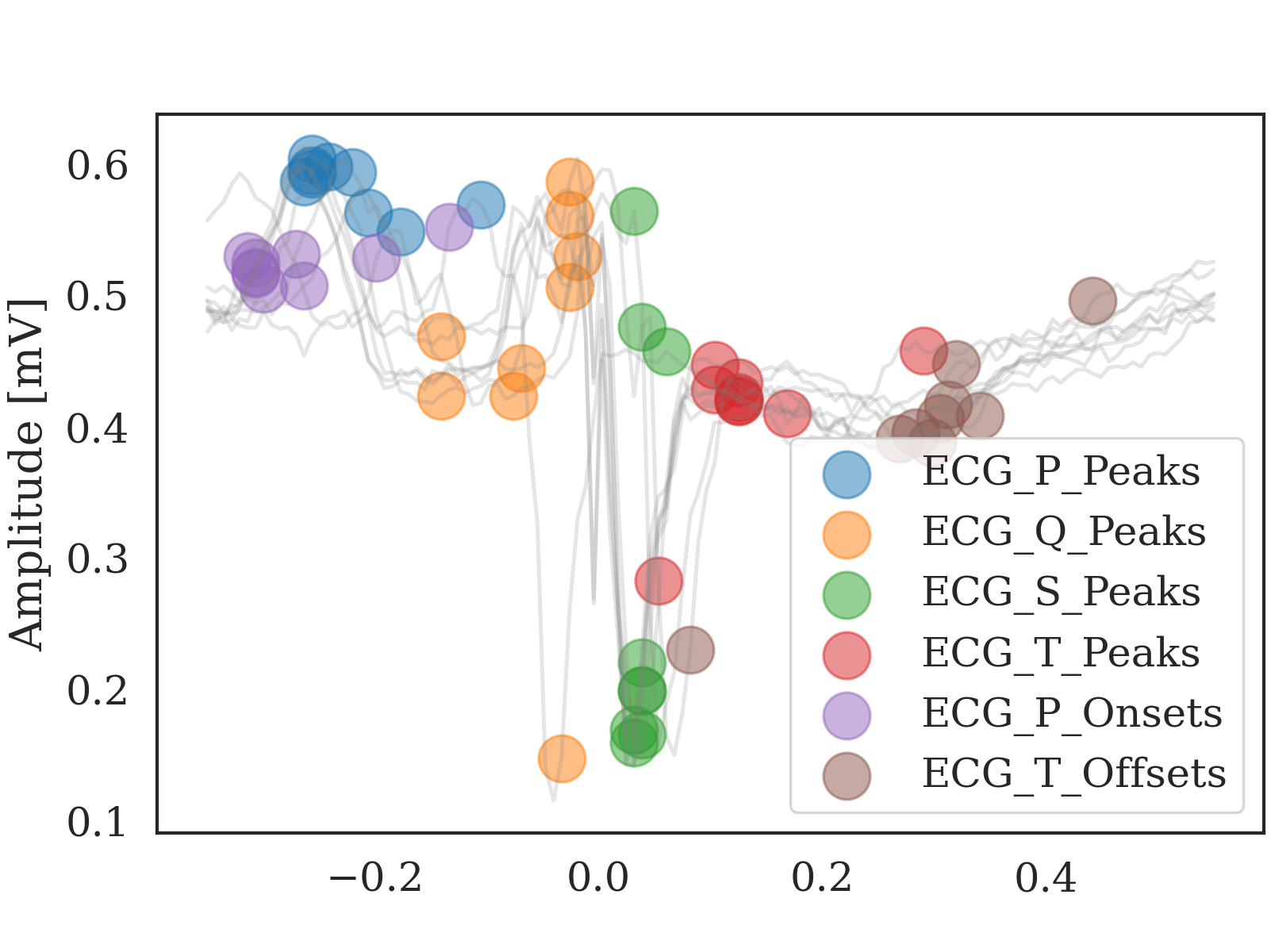}\label{fig:morphe-a}}
\hspace{30pt}
\subfloat[]{\includegraphics[width=0.35\textwidth]{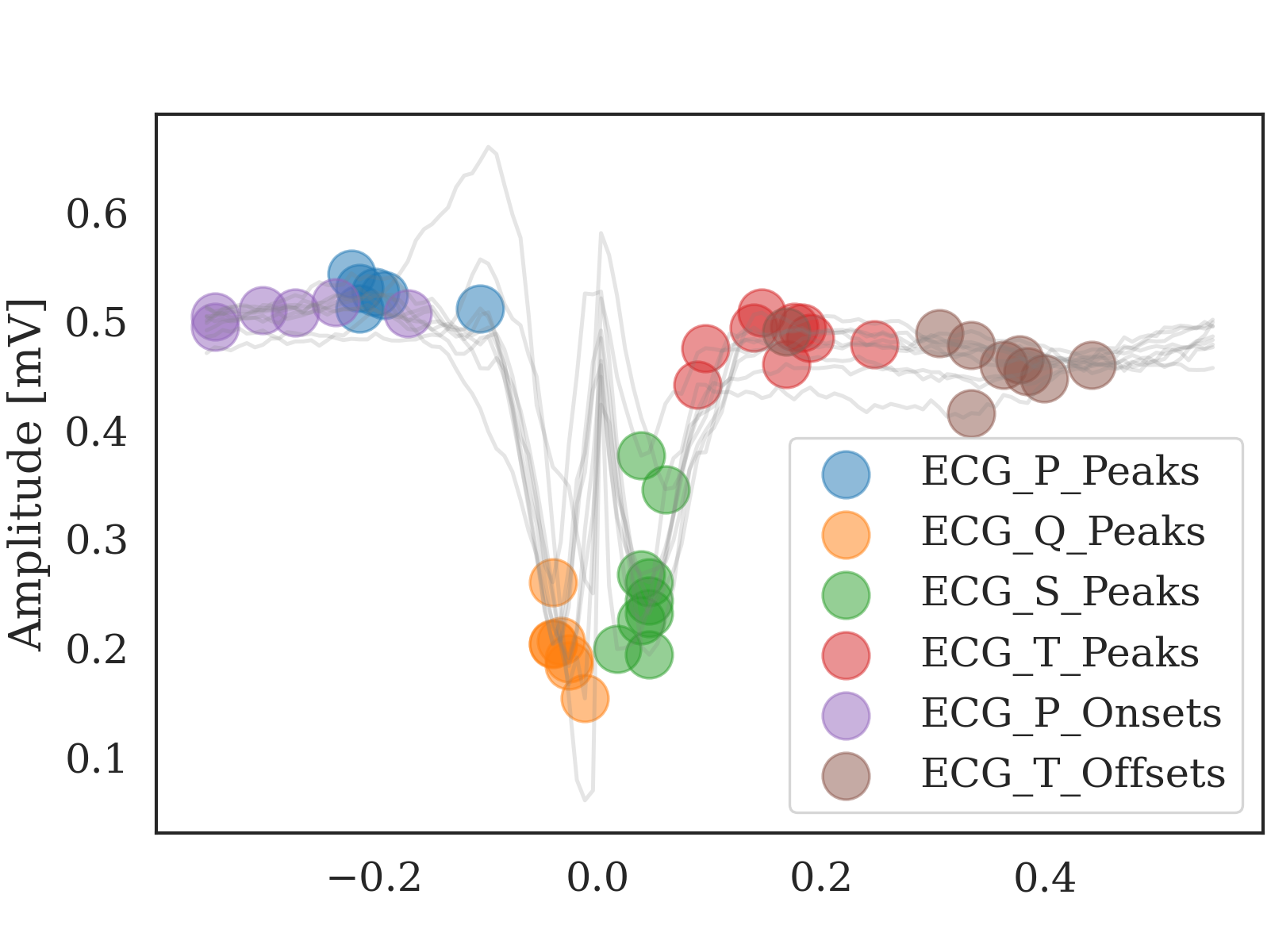}\label{fig:morphe-b}}
\caption{Visual summary of the segmentation of P,Q,S and T peak onsets and offsets for both normal and anomalous classes (\cite{neurokit}).}
\label{fig:peaks}
\end{figure}

As a further quantitative evidence of the overall adherence of the generated signals to the corresponding ECG classes from the original datasets, we retrieved the period of meaningful segments. The latter have been carried out via an automatic feature extraction phase (Figure \ref{fig:peaks}), with the objective of measuring: QRS complexes, QT, ST and PR segments. 

Distinctive traits and patterns of PVCs, differently from normal rhythms, emerge also for the generated signals as can be observed in Table \ref{tab:segments}. These properties are clearly desirable both for patterns fidelity and data augmentation procedures involving classifiers relying on those features.  

\begin{table}[!htb]
\centering
\caption{Summary of the time duration of meaningful segments after feature extracting Q,R,S and T peaks. A comparison between the distribution of segments duration in seconds between real BIDMC samples those produced by ECGAN.}
  {\begin{tabular}{p{\dimexpr 0.2\linewidth-2\tabcolsep} 
                   p{\dimexpr 0.2\linewidth-2\tabcolsep} 
                   p{\dimexpr 0.2\linewidth-2\tabcolsep} 
                   p{\dimexpr 0.2\linewidth-2\tabcolsep} 
                   p{\dimexpr 0.2\linewidth-2\tabcolsep}}
  \toprule
  Dataset class & QT (s) & QRS (s) & PR(s) & ST(s) \\
  \midrule
    BIDMC (N)	 &	 $ 0.55 \pm 0.13$	 &	 $ 0.11 \pm 0.03$ &	 $ 0.60 \pm 0.12$	 &	 $ 0.24 \pm 0.05$	\\	
	 BIDMC (V)	 &	 $ 0.52 \pm 0.11$	 &	 $ 0.06 \pm 0.04$ &	 $ 0.61 \pm 0.03$	 &	 $ 0.13 \pm 0.09$ \\
	 \midrule
	 ECGAN (N)	 &	 $ 0.53 \pm 0.15$	 &	 $ 0.13 \pm 0.05$	 &	 $ 0.59 \pm 0.07$	 &	 $ 0.26 \pm 0.08$	\\
	 ECGAN (V)	 &	 $ 0.60 \pm 0.05$	 &	 $ 0.12 \pm 0.11$ &	 $ 0.60 \pm 0.05$	 &	 $ 0.12 \pm 0.11$	\\
\bottomrule
\label{tab:segments}
\end{tabular}}
\end{table}

\begin{figure}[!htb]
\centering
\caption{Morphological distribution of a normal ECG (left) and anomalous (right). The top row is sampled from BIDMC, while the bottom is generated by the proposed model.}
\includegraphics[width=0.6\linewidth]{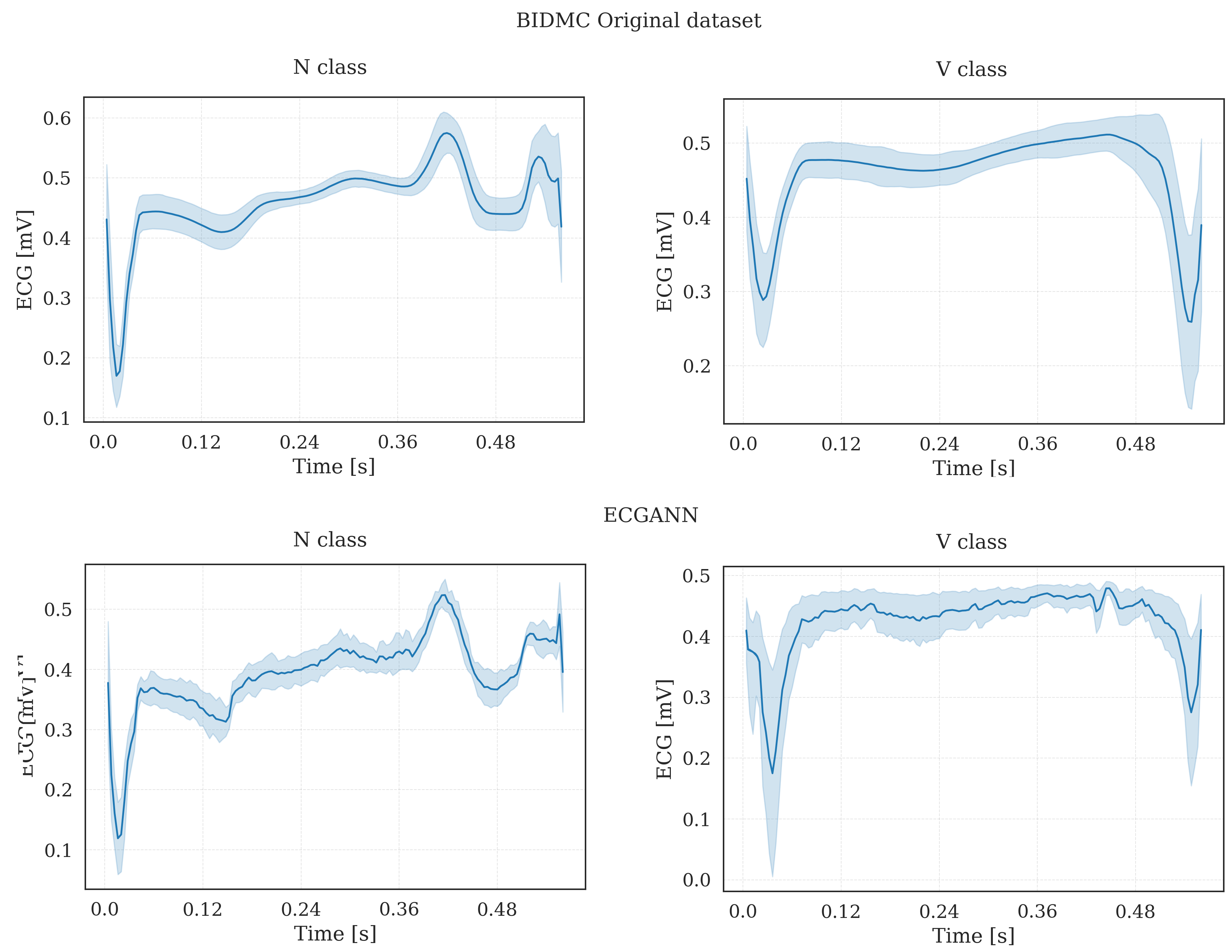}
\label{fig:morph}
\end{figure}

We proceed by showing the morphological distribution of a subset of synthesized and baseline ECG patterns from the first dataset (Figure \ref{fig:morph}). The overall adherence to segment lengths is respected without sacrificing sampling diversity.


\section{Functionality assessment}\label{apd:funct}

The architecture of inner feature learners is flexible allowing us to utilize stacked LSTMs for an higher level of abstraction. Meanwhile, the discriminator is built upon three stacked convolutional blocks: 1D convolutional layer followed by global average pooling and instance normalization. The latter allowed significant improvements over batch normalization. The choice of adopting LSTMs for our generative task is based on the requirement of learning long-term sequential rather than spatial features. Differently, convolutional blocks are better suited for our discriminative duty by focusing on local patterns in parallel. The details of these architectural choices are reported extensively in Table \ref{tab:architectural}.

\begin{table}[!h]
\centering
\caption{Architectural details of the layers from the discriminator and the generator.}
\begin{tabular}{p{\dimexpr 0.5\linewidth-2\tabcolsep} 
                   p{\dimexpr 0.5\linewidth-2\tabcolsep}}
\toprule
\multicolumn{2}{c}{Generator} \\
\midrule
Component & Shape \\
\midrule
LSTM                 &    $128 \times h$ \\
LSTM ($\times 4$)            &    $128 \times n$ \\
\midrule
\multicolumn{2}{c}{Discriminator} \\
\midrule
Component & Shape \\
\midrule
1D CNN (kernel size = 6) & $128 \times n$ \\
Instance Normalization & $128 \times n$ \\
ReLU Activation & $128 \times n$ \\
1D CNN (kernel size = 6) & $128 \times n$ \\
Instance Normalization & $64 \times n$ \\
ReLU Activation & $64 \times n$ \\
1D CNN (kernel size = 6) & $32 \times n$ \\
Instance normalization & $32 \times n$ \\
ReLU Activation & $32 \times n$ \\ 
Global average pooling  & $32 \times 1$ \\
Linear activation  & $1 \times 1$ \\ 
\bottomrule
\label{tab:architectural}
\end{tabular}
\end{table}

The architecture of the CNN-based arrhythmia classifier $\mathcal{C}$, have been employed also to evaluate generated signals from a functional perspective. The objective of this phase is exploiting the inclusion of synthetic data to the training set to improve generalization capabilities on the initially retained internal test set. In the long run, each of the baselines achieved improved performances except for audio synthesis models (SpecGAN and WaveGAN).

\begin{table}[!h]
\centering
\caption{Predictive results for the classification task.  A fixed number of balanced samples for each of the generative models is randomly inserted within the training set, and later evaluated over the test set.}
  {\begin{tabular}{p{\dimexpr 0.17\linewidth-2\tabcolsep} 
                   p{\dimexpr 0.17\linewidth-2\tabcolsep} 
                   p{\dimexpr 0.17\linewidth-2\tabcolsep} 
                   p{\dimexpr 0.17\linewidth-2\tabcolsep} 
                   p{\dimexpr 0.17\linewidth-2\tabcolsep}
                   p{\dimexpr 0.17\linewidth-2\tabcolsep}}
  \toprule
  Dataset & Accuracy & Specificity & Sensitivity & Precision & F1\\
  \midrule
  \multicolumn{6}{c}{Training}\\
  \midrule
    Original 	 &	 $ 0.990 \pm 0.01$	 &	 $ 0.986 \pm 0.02$	 &	 $ 0.996 \pm 0.04$	 &	 $ 0.997 \pm 0.03$	 &	 $ 0.991 \pm 0.02$	\\
    ECGAN 	 &	 $ 0.997 \pm 0.01$	 &	 $ 0.999 \pm 0.01$	 &	 $ 0.994 \pm 0.03$	 &	 $ 0.996 \pm 0.02$	 &	 $ 0.997 \pm 0.01$	\\
    TimeGAN 	 &	 $ 0.987 \pm 0.02$	 &	 $ 0.980 \pm 0.03$	 &	 $ 0.997 \pm 0.03$	 &	 $ 0.997 \pm 0.02$	 &	 $ 0.988 \pm 0.01$	\\
    C-RNNGAN 	 &	 $ 0.991 \pm 0.01$	 &	 $ 0.985 \pm 0.02$	 &	 $ 0.999 \pm 0.01$	 &	 $ 0.999 \pm 0.00$	 &	 $ 0.992 \pm 0.01$	\\
    SpecGAN 	 &	 $ 0.960 \pm 0.03$	 &	 $ 0.931 \pm 0.05$	 &	 $ 0.998 \pm 0.01$	 &	 $ 0.998 \pm 0.01$	 &	 $ 0.963 \pm 0.03$	\\
    WaveGAN 	 &	 $ 0.937 \pm 0.04$	 &	 $ 0.925 \pm 0.08$	 &	 $ 0.954 \pm 0.07$	 &	 $ 0.969 \pm 0.04$	 &	 $ 0.943 \pm 0.03$	\\
	 \midrule
  \multicolumn{6}{c}{Testing}\\
  \midrule
	 Original  &	 $ 0.981 \pm 0.02$	 &	 $ 0.988 \pm 0.02$	 &	 $ 0.973 \pm 0.09$	 &	 $ 0.979 \pm 0.07$	 &	 $ 0.991 \pm 0.02$	\\
	 ECGAN 	 &	 $ 0.989 \pm 0.01$	 &	 $ 0.998 \pm 0.00$	 &	 $ 0.977 \pm 0.03$	 &	 $ 0.982 \pm 0.02$	 &	 $ 0.997 \pm 0.01$	\\
	  
	 TimeGAN 	 &	 $ 0.988 \pm 0.01$	 &	 $ 0.995 \pm 0.04$	 &	 $ 0.978 \pm 0.05$	 &	 $ 0.983 \pm 0.04$	 &	 $ 0.988 \pm 0.01$ \\
	 
	 C-RNNGAN 	 &	 $ 0.987 \pm 0.01$	 &	 $ 0.994 \pm 0.00$	 &	 $ 0.979 \pm 0.05$	 &	 $ 0.984 \pm 0.04$	 &	 $ 0.992 \pm 0.01$	\\
	 
	 SpecGAN 	 &	 $ 0.969 \pm 0.02$	 &	 $ 0.956 \pm 0.05$	 &	 $ 0.985 \pm 0.00$	 &	 $ 0.988 \pm 0.00$	 &	 $ 0.963 \pm 0.03$	\\
	 
	 WaveGAN 	 &	 $ 0.950 \pm 0.03$	 &	 $ 0.927 \pm 0.07$	 &	 $ 0.979 \pm 0.02$	 &	 $ 0.983 \pm 0.01$	 &	 $ 0.943 \pm 0.04$	\\
	 
\bottomrule
\label{tab:ext_pred_res}
\end{tabular}}
\end{table}
\end{document}